\documentclass{article}
\pdfoutput=1
\usepackage{spconf,amsmath,graphicx}
\usepackage{mathrsfs,cite}
\usepackage{paralist}
\usepackage{booktabs}
\usepackage{microtype}

\usepackage{xcolor}

\title{LEVERAGING UNPAIRED TEXT DATA FOR TRAINING END-TO-END SPEECH-TO-INTENT SYSTEMS}
\name{\begin{tabular}{c}Yinghui Huang, Hong-Kwang Kuo, Samuel Thomas, Zvi Kons$^{\dagger}$\\
Kartik Audhkhasi, Brian Kingsbury, Ron Hoory$^{\dagger}$, Michael Picheny\sthanks{The author performed the work while at IBM}\end{tabular}}
\address{IBM Research AI, Yorktown Heights, USA\\
    $^{\dagger}$IBM Research AI, Haifa, Israel}
\begin{document}
\ninept
\maketitle
\begin{abstract}

Training an end-to-end (E2E) neural network speech-to-intent (S2I)
system that directly extracts intents from speech requires
large amounts of intent-labeled speech data, which is time consuming and
expensive to collect. Initializing the S2I model with an ASR model
trained on copious speech data can alleviate data
sparsity. In this paper, we attempt to leverage NLU text resources. We implemented a CTC-based S2I system that matches the
performance of a state-of-the-art, traditional cascaded SLU system.  We performed controlled experiments with varying
amounts of speech and text training data.  When only a tenth of the
original data is available, intent classification accuracy degrades by
7.6\% absolute.  Assuming we have additional text-to-intent data
(without speech) available, we investigated two techniques to improve
the S2I system:
\begin{inparaenum}[(1)]
\item transfer learning, in which acoustic
embeddings for intent classification are tied to fine-tuned BERT text embeddings; and
\item data augmentation, in which the text-to-intent data is converted into
speech-to-intent data using a multi-speaker text-to-speech system.
\end{inparaenum}
The proposed approaches recover 80\% of performance lost due to using limited intent-labeled speech.
\end{abstract}
\section{Introduction}
\label{sec:intro}

Spoken language understanding (SLU) systems have traditionally been a cascade of an automatic speech recognition (ASR) system converting speech into text followed by a natural language understanding (NLU) system that interprets the meaning, or intent, of the text.  In contrast, an end-to-end (E2E) SLU system~\cite{serdyuk2018towards,qian2017exploring,chen2018spoken,ghannay2018end,lugosch2019speech,Haghani2018,caubriere2019curriculum} processes speech input directly into intent without going through an intermediate text transcript.
There are many advantages of end-to-end SLU systems \cite{lugosch2019speech}, the most significant of which is that E2E systems can directly optimize the end goal of intent recognition, without having to perform intermediate tasks like ASR.  %

Compared to end-to-end SLU systems, cascaded systems are modular and each component can be optimized separately or jointly (also with end-to-end criteria~\cite{Goel2005,yaman2008integrative,Haghani2018}).  One key advantage of modular components is that each component can be trained on data that may be more abundant.  For example, there is a lot of transcribed speech data that can be used to train an ASR model.  In comparison, there is a paucity of speech data with intent labels, and intent labels, unlike words, are not standardized and may be inconsistent from task to task.  Another advantage of modularity is that components can be re-used and adapted for other purposes, e.g. an ASR service used as a component for call center analytics, closed captioning, spoken foreign language translation, etc.

While end-to-end SLU is an active area of research, currently the most promising results under-perform or just barely outperform traditional cascaded systems~\cite{caubriere2019curriculum,qian2018speech}.  One reason is that deep learning models require a large amount of appropriate training data.  To train an end-to-end speech-to-intent model, we need intent-labeled speech data, and such data is usually scarce. \cite{caubriere2019curriculum,tomashenko2019investigating} address this problem using a curriculum and transfer learning approach whereby the model is gradually trained on increasingly relevant data until it is fine-tuned on the actual domain data.  Similarly,~\cite{lugosch2019speech,bhosale2019end} advocate pre-training an ASR model on a large amount of transcribed speech data to initialize a speech-to-intent model that is then trained on a much smaller training set with both transcripts and intent labels.

Training data for end-to-end SLU is much scarcer than training data for ASR (speech and transcripts) or NLU (text and semantic annotations). In fact, there are many relevant NLU text resources and models (e.g. named entity extraction) and information in the world is mostly organized in text format, without corresponding speech.  As SLU becomes more sophisticated, it is important to be able to leverage such text resources in end-to-end SLU models. Pre-training on ASR resources is straightforward, but it is less clear how to use NLU resources. This problem has not been adequately addressed in the literature that we are aware of.

In this paper, we pose an interesting question: for an end-to-end S2I model, how can we take advantage of text-to-intent training data without speech?   There are many possible approaches, but we focus on two methods in this paper.  In the first, we jointly train the speech-to-intent model and a text-to-intent model, encouraging the acoustic embedding from the speech model to be close to a fine-tuned BERT-based text embedding, and using a shared intent classification layer. The second method involves data augmentation,
where we convert the additional text-to-intent data into synthetic speech-to-intent data using a multi-speaker text-to-speech (TTS) system.

To evaluate our methods, we performed carefully controlled experiments.  First we built 
 strong baselines with conventional cascaded systems, where the acoustic, language, and intent classification models are adapted on in-domain data.  We also built a strong speech-to-intent model using pre-trained acoustic models for initialization and multi-task training to optimize ASR and intent classification objectives, producing competitive results compared with the cascaded system.  We evaluated these models on varying amounts of speech and text training data.  Through these experiments, we seek to answer two questions. First, can we improve S2I models with additional text-to-intent data? Second, how does that compare to having actual speech-to-intent data? 

\section{Training Speech-to-Intent Systems}
\label{sec:s2i}
End-to-end speech-to-intent systems directly extract the intent label associated with a spoken utterance without explicitly transcribing the utterance. However, it is still useful to derive an intermediate ASR embedding that summarizes the message component of the signal for intent classification. An effective approach to achieve this goal is to train the S2I classifier starting from a pre-trained ASR system. 
ASR pre-training is also beneficial since intent labels are not required in this step; hence, we can use ASR speech data instead of specific in-domain intent data, which is usually limited. 

  \begin{figure}[t]
    \begin{center}
        \includegraphics[width=6cm]{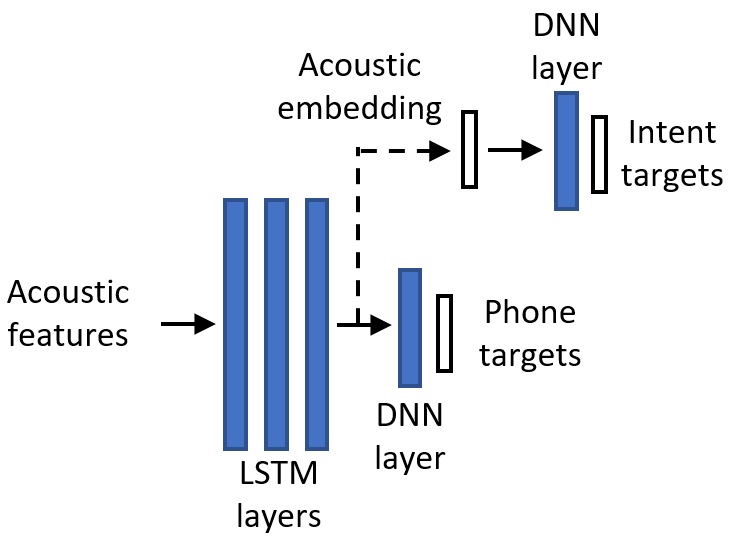}
    \end{center}
  \vspace{-4mm}
    \caption{A S2I system with pre-trained ASR}
    \label{fig:asr_pretrain}     
  \end{figure}

In our work, we use a phone-based connectionist temporal classification (CTC)~\cite{graves2006connectionist} acoustic model (AM) trained on general speech data as the base ASR system. First, we initialize the S2I model with this model and adapted it to the in-domain data. Once the adapted ASR system is trained, it is modified for intent classification using speech that was transcribed and also annotated with intent labels. As shown in Fig.~\ref{fig:asr_pretrain}, to construct the intent recognition system we modify the model by adding a classification layer that predicts intent targets. Unlike phone targets which are predicted by the ASR system at the frame level, intent targets span larger contexts. In our case, we assume that each training utterance corresponds to a {\em single}\/ intent, although it might be composed of several words. To better capture intents at the utterance level, we  derive an acoustic embedding (AE) corresponding to each training utterance. This embedding is computed by time averaging all the hidden states of the final LSTM layer to summarize the utterance into one compact vector that is used to predict the final intent. The final fully connected layer introduced in this step to process the acoustic embeddings can be viewed as an intent classifier. While training the network to predict intent, given that transcripts for the utterances are also available, we continue to refine the network to predict ASR targets as well. With this multi-task objective, the network adapts its layers to the channel and speakers of the in-domain data.
 During test time, only the outputs of the intent classification layer are used, while the ASR branch is discarded. To improve robustness of the underlying speech model, we also employ data augmentation techniques for end-to-end acoustic model training, namely speed and tempo perturbation, in all our experiments.

\section{Improvements using Text-to-Intent data}
\label{sec:limited_data}
In practice, we expect that end-to-end S2I classifiers will be trained in conditions where there is a limited amount of transcribed S2I data and significantly more text-to-intent (T2I) data.  To investigate this setting, we develop two approaches to use additional text data for building S2I systems. 

\subsection{Leveraging pre-trained text embedding}
\label{sec:bert}

\begin{figure}[t]
    \begin{center}
        \includegraphics[width=7.5cm]{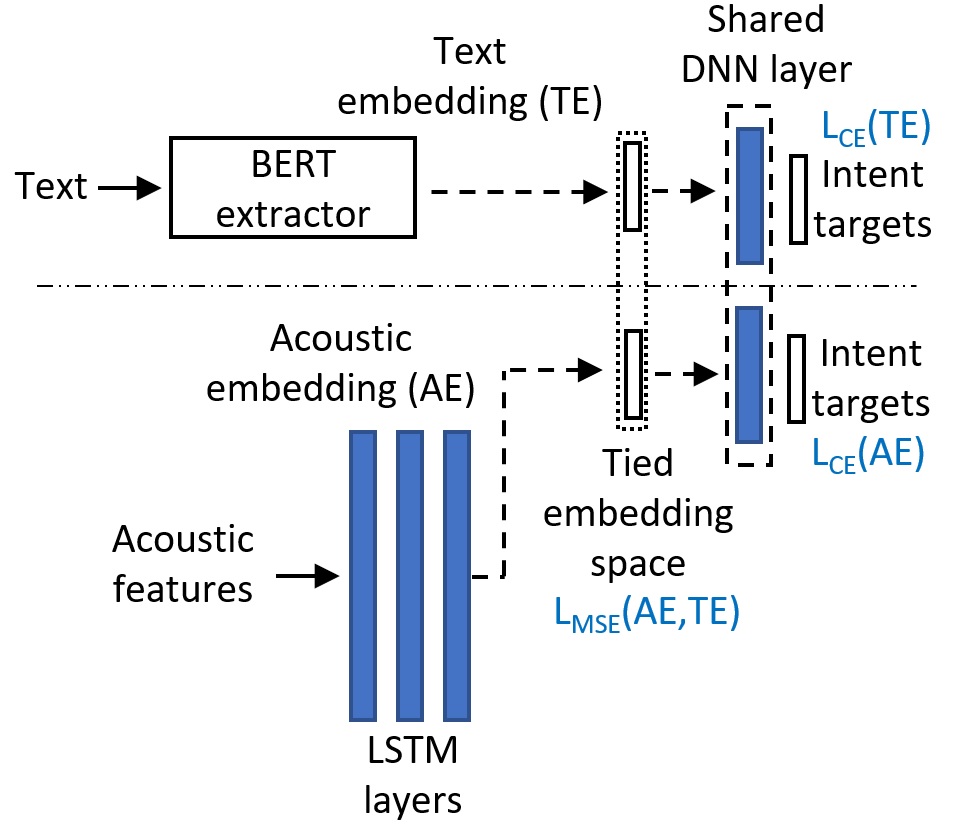}
    \end{center}
    \vspace{-3mm}
    \caption{Joint-training of the S2I system with text embeddings}
    \label{fig:joint-train}
    \vspace{-5mm}
  \end{figure}

Leveraging text embedding (TE) from models pre-trained on large amounts of data, such as BERT~\cite{devlin2018bert} and GPT-2~\cite{radford2019language}, has recently improved performance in a number of NLU tasks. In this paper, we use BERT-based text embeddings to transfer knowledge from text data into a speech-to-intent system. The text embeddings are used to ``guide'' acoustic embeddings which are trained with a limited amount of S2I data, in the same spirit as learning a shared representation between modalities~\cite{ngiam2011multimodal,andrew2013deep,wang2015deep,harwath2018jointly}.
We employ the following steps to train the final model. \\
{\bf (A) T2I model pre-training:} As in the standard process outlined in the original BERT paper~\cite{devlin2018bert}, we first fine-tune BERT on the available text-to-intent data using a masked language model (LM) task as the intermediate task. The model is further fine-tuned with intent labels as the target classification task before the representation of the special token \texttt{[CLS]} is used as the text embedding of an utterance. \\
{\bf (B) ASR pre-training for S2I model:} As described in Section~\ref{sec:s2i}, the base ASR model is trained on non-parallel ASR data and subsequently adapted using an augmented S2I data set. This step adapts the model to the acoustic conditions of the intent data to extract better acoustic embeddings. Next, multi-task training further fine-tunes the acoustic model to generate embeddings that are better for intent classification. We now have an initial S2I system that has seen only limited novel intent content. Note that we add a fully connected layer before the classifier to make sure the dimensions of the acoustic embedding and text embedding match for joint-training in the next step. \\
{\bf (C) Full S2I model training:} The final S2I classifier is then assembled by combining the fine-tuned T2I classifier (step A) and the pre-trained S2I system (step B). We jointly train the fine-tuned BERT model with the pre-trained ASR acoustic model in an attempt to leverage the knowledge extracted from larger amounts of text data to improve the quality of the acoustic embedding for intent classification. The training framework is shown in Figure~\ref{fig:joint-train}. We extract text embeddings from the fine tuned BERT model with the reference text as input. Acoustic embeddings are also extracted in parallel from a corresponding acoustic signal. These two embeddings are used to train a shared intent classification layer which has been initialized from the  text-only classification task described above. Given that the text embedding comes from a well trained extractor, the acoustic embeddings are explicitly forced to match the better text embeddings. We hypothesize that this matching will also allow the shared classifier layer to train better. During test time, we only use the acoustic branch for intent inference. 

To achieve these goals, a training procedure that optimizes two separate loss terms is employed. %
The first loss term corresponds to a composite cross-entropy intent classification loss derived by using the text embeddings, ${L}_{CE}(TE)$, and the acoustic embeddings, ${L}_{CE}(AE)$, separately to predict intent labels using the shared classifier layer. In the combined classification loss, the text-embedding classification loss is scaled by a weight parameter $\alpha$. 
The second loss is the mean squared error (MSE) loss between the text embedding and acoustic embedding ${L}_{MSE}(AE,TE)$. It is important to note that while the gradients from the combined classification loss are propagated back to both the text and acoustic embedding networks, the MSE loss is only back-propagated to the acoustic side because we presume that the acoustic embeddings should correspond closely to the BERT embeddings, which have been trained on massive quantities of text and perform better on intent classification. On the speech branch the minimized loss is \mbox{${L}_{MSE}(AE,TE)+{L}_{CE}(AE)+\alpha{L}_{CE}(TE)$}, while the loss on the text branch is \mbox{${L}_{CE}(AE)+\alpha{L}_{CE}(TE)$}.

\subsection{Using text data for speech data augmentation}

Instead of using available text data for pre-training the T2I system, we also try converting the text data to speech using a multi-speaker TTS system. %
Like in Section~\ref{sec:s2i}, the S2I classifier in this case is trained in two steps. \\
{\bf (A) ASR pre-training:} The base ASR model (trained on non-parallel ASR data) is first adapted using the speech portion of the S2I data set. \\
{\bf (B) Training with synthesized data:} The TTS-synthesized data is used along with the limited amount of S2I data, without augmentation, for training. 
Section~\ref{sec:tts} describes the generation of the TTS data set in detail. 
Compared with tying text and acoustic embeddings, S2I data augmentation might be more effective at training the ``embedding'' to intent classifier layers of the neural network, because novel (utterance, intent) pairs are used in the final training phase rather than just in pre-training.

\section{Experiments and Results}
\label{sec:expt}

\subsection{Experimental setup}

Experiments were performed on a corpus consisting of call center recordings of open-ended first
utterances by customers describing the reasons for their calls, which is described in~\cite{Goel2005}.  The
8kHz telephony speech data was manually transcribed and labeled with
correct intents.  The corpus contains real customer spontaneous utterances, 
not crowd-sourced data of people reading from a script, and includes a variety 
of ways customers naturally described their intent.  For example, 
the intent ``BILLING'' includes short sentences such as ``{\em billing}'' 
and longer ones such as ``{\em i need to ask some questions on uh to get 
credit on an invoice adjustment}.''

The training data consists of 19.5 hours of speech that was first divided
into a training set of 17.5 hours and a held-out set of 2 hours.  The held-out set
was used during training to track the objective function and
tune certain parameters like the initial learning rate.  In addition to
the 17.5-hour training set (which we call {\em 20hTrainset}, containing 21849
sentences, 145K words), we also extracted a 10\% subset (1.7h) for
low-resource experiments (which we call {\em 2hTrainset}, containing 2184
sentences, 14K words).  We augmented the training data via speed and tempo perturbation (0.9x and 1.1x), so {\em 2hTrainset}\/ finally contains about 8.7 hours and {\em 20hTrainset}\/ about 88 hours of speech. The {\em devset}\/ consists of 3182 sentences (2.8 hours) and was used for hyperparameter tuning, e.g., tuning the acoustic weight to optimize the word error rate (WER).  A
separate data set containing 5592 sentences (5h, 40K words) was used
as the final {\em testset}.  

In the training set, each sentence had a single intent, and there were
29 intent classes.  The testset contains additional unseen
intent classes and multiple intents per sentence, as naturally happens
in real life.  For simplicity, in this paper we do not address these
cases and always count such sentences as errors when calculating
intent accuracy; they account for about 1\% of the utterances.  The {\it testset}\/ has an
average of 7 words per utterance, with the longest sentence being over 100
words long. 70\% of the sentences are unique (not repetitions).

\subsection{Pre-trained models}
\subsubsection{ASR CTC model}
\label{sec:asr_sys}
We pre-train an ASR system on the 300-hour Switchboard English conversational speech corpus. The AM is a 6-layer bidirectional LSTM network with every layer containing 640 LSTM units per direction. The AM is trained using CTC loss over 44 phones and the blank symbol. Our AM training recipe follows our prior work. We first perform speed and tempo perturbation (0.9x and 1.1x) resulting in a 1500-hour audio data set. We train the AM for 20 epochs using CTC loss, followed by 20 epochs of soft forgetting training~\cite{audhkhasi2019forget}, followed by 20 epochs of guided training~\cite{kurata2019guiding}. We use sequence noise injection~\cite{saon2019sequence} and SpecAugment~\cite{park2019specaugment} throughout the training to provide on-the-fly data augmentation, and we also use a dropout probability of 0.5 on the LSTM output at each layer.  %

\subsubsection{BERT based T2I model}
We start with pre-trained BERT\textsubscript{base} model of~\cite{devlin2018bert}. Using the implementation introduced in \cite{wolf2019transformers}, we first pre-train using a masked LM target with learning rate $3\text{e-}5$ for 10 epochs, followed by 3 epochs of fine-tuning on the intent classification task with learning rate $2\text{e-}5$. This text-to-intent BERT based classifier trained on {\em 20hTrainset}\/ gives 92.0\% accuracy on human-generated reference transcripts. 

\subsubsection{TTS system}
\label{sec:tts}
The TTS system architecture is similar to the single speaker system described in \cite{Kons2019}. It is a modular system based on three neural-net models: one to infer prosody, one to infer acoustic features, and an LPCNet~\cite{Valin2019LPCNet} vocoder.
The main difference between the single and multi-speaker systems is that both the prosody and the acoustic networks are converted to multi-speaker models by conditioning them on a speaker embedding vector. Each of the three models was independently trained on 124 hours of 16KHz speech from 163 English speakers. The speaker set is composed of 4 high quality proprietary voices with more than 10 hours of speech, 21 VCTK~\cite{VCTK2017} voices, and 138 LibriTTS~\cite{Zen2019LibriTTSAC} voices. During synthesis, for each sentence we select a random speaker out of the known speakers set. We then synthesize the sentence with this voice. Finally, the samples are downsampled to 8KHz to match the S2I audio sampling rate.

\subsection{Results}

We first establish the strongest possible baseline results for the conventional cascaded system where the ASR and T2I models are trained separately. %
Using the ASR system described above and a BERT T2I model trained on the domain data {\it 20hTrainset}, we obtained 21.9\% WER and 73.4\% intent classification accuracy (IntAcc) for {\it testset}, as shown in Table~\ref{tab:baseline}.  %
The same T2I model has 92.0\% intent accuracy on human reference transcripts; thus, there is a significant degradation in accuracy with speech input due to ASR errors. %
Both the AM and LM of this system are then adapted using the domain data {\it 20hTrainset}.  Table~\ref{tab:baseline} shows that such adaptation 
dramatically improved both WER and intent accuracy, which increased from 73.4\% to 89.7\%.  There is now only about a 2\% accuracy gap between using human transcripts (92.0\%) and using ASR outputs (89.7\%).  In the low-resource scenario, adapting the AM and LM on {\it 2hTrainset} and also training T2I on only {\it 2hTrainset} results in intent accuracy of 82.8\%.

\begin{table}%
	\centering
	\begin{tabular}{@{}lllcc@{}} \toprule
	\multicolumn{1}{c}{\bf AM} & \multicolumn{1}{c}{\bf LM} & \multicolumn{1}{c}{\bf T2I} & {\bf WER} & \bf{IntAcc} \\ \cmidrule(lr){1-3}
    \cmidrule(lr){4-5}
		unadapted   & unadapted   & 20hTrainset & 21.9\%  & 73.4\% \\
		20hTrainset & 20hTrainset & 20hTrainset & 10.5\% & 89.7\% \\ 
		2hTrainset & 2hTrainset & 2hTrainset  & 11.6\% & 82.8\% \\ \bottomrule
	\end{tabular}
	\caption{WER and IntAcc (intent accuracy) of baseline cascaded systems (ASR followed by T2I) for speech-to-intent recognition}
	\label{tab:baseline}
\end{table}

As shown in Table~\ref{tab:E2E}, when paired speech-to-intent data is used our proposed E2E S2I approach gives comparable accuracy as the cascaded approach, both with full training data ({\it 20hTrainset}) or in the low-resource setting ({\it 2hTrainset}). 

In the low-resource scenario where only a limited amount of speech is available ({\em 2hTrainset}), frequently one may have extra text data with intent labels but no corresponding speech data.  For the cascaded system, it is straightforward to train different components (AM, LM, T2I) with whatever appropriate data is available.   %
Table~\ref{tab:extraText} shows how the intent accuracy varies when the full {\it 20hTrainset} (but not the speech) is available as text-to-intent data. By training the LM and T2I on this data, the intent accuracy increased to 89.6\%, basically matching the best accuracy of 89.7\%, where the AM is also adapted on {\it 20hTrainset}. The third row in Table~\ref{tab:extraText} shows that if only the T2I model is trained on {\it 20hTrainset}, the accuracy is still quite high: 88.9\%.  Comparing results from the first and third rows, we observe that the accuracy difference between training the T2I model on {\it 2hTrainset}\/ versus {\it 20hTrainset}\/ is about 6\%, accounting for most of the gap between the full resource and limited resource performance.  In other words, the T2I model is the weakest link, the component that is most starved for additional data.   For the AM and LM, because they were pre-trained on plenty of general data, even just 2h of adaptation is almost enough, but the intent classifier needs much more data.  This makes sense because the intent classification task can be quite domain specific.

\begin{table}%
	\centering 
	\begin{tabular}{@{}lcc@{}} \toprule
		 & \multicolumn{1}{c}{\bf 20hTrainset} & \multicolumn{1}{c}{\bf 2hTrainset}  \\ \midrule
		Cascaded(ASR+T2I) & 89.7\% & 82.8\% \\ 
		E2E CTC & 89.8\% & 82.2\% \\ 
		\bottomrule
	\end{tabular}
	\caption{End-to-end speech-to-intent classification accuracy.}
	\label{tab:E2E}
\end{table}

\begin{table}%
	\centering 
	\begin{tabular}{@{}lllcc@{}} 
	\toprule
		\multicolumn{1}{c}{\bf AM} & \multicolumn{1}{c}{\bf LM} & \multicolumn{1}{c}{\bf T2I} & {\bf WER} & \bf{IntAcc} \\ \cmidrule(lr){1-3}
        \cmidrule(lr){4-5}
		2hTrainset & 2hTrainset & 2hTrainset   & 11.6\% & 82.8\% \\ 
		2hTrainset & 20hTrainset & 20hTrainset & 10.3\% & 89.6\% \\ 
		2hTrainset & 2hTrainset & 20hTrainset  & 11.6\% & 88.9\% \\ 
		\bottomrule
	\end{tabular}
	\caption{Limited speech resources with extra text-to-intent data.}
	\label{tab:extraText}
\end{table}

For the end-to-end speech-to-intent system, leveraging the text-to-intent data is not straightforward.  If it were unable to take advantage of such data, it would be at a significant 6-7\% accuracy disadvantage compared to the cascaded system in this scenario.  

\begin{table}%
	\centering 
	\begin{tabular}{@{}lc@{}} 
	    \toprule
		\multicolumn{1}{c}{\bf Method} &  \multicolumn{1}{c}{\bf IntAcc} \\ \midrule
		E2E S2I system trained on 2hTrainset & 82.2\% \\ \midrule
		Joint training tying speech/text embeddings & 84.7\% \\ 
		Adding synthetic multi-speaker TTS speech & 87.8\% \\ 
		Joint training + adding synthetic speech  & 88.3\% \\  \midrule
		E2E S2I system trained on 20hTrainset & 89.8\% \\ 
		\bottomrule
	\end{tabular}
	\caption{End-to-End models using extra text-to-intent data to recover accuracy lost by switching from {\em 20hTrainset}\/ to {\em 2hTrainset}.}
	\label{tab:extraTextE2E}
\end{table}

In the next set of experiments, in Table~\ref{tab:extraTextE2E}, we show results from leveraging extra text data to reduce the gap caused by having less S2I training data. 
Our first approach described in Section~\ref{sec:bert} and illustrated in Figure~\ref{fig:joint-train} ties the speech and text embeddings and trains the intent classifier on speech and text embeddings.  By joint training end-to-end S2I CTC model with BERT fine tuned on full text-to-intent data, we observe accuracy improvement from 82.2\% to 84.7\%. 

In our second approach,  we took the extra text-to-intent data ({\em 20hTrainset}) and generated synthetic speech with multi-speaker TTS to create new speech-to-intent data.  Adding this synthetic data to {\em 2hTrainset} to train the E2E model resulted in an intent accuracy of 87.8\%, a significant improvement from 82.2\%.  If we generated the synthetic data using single speaker TTS, the accuracy was roughly the same: 87.3\%.  These results were quite surprising.  %
We hypothesize that this improvement is due to the S2I model learning new semantic information (``embedding''-to-intent) from the new synthetic data rather than adapting to the acoustics. Therefore it was not necessary to generate a lot of variability in the speech (e.g. speakers, etc.) with the TTS data.  Running ASR on the TTS speech, the WER was very low, around 4\%, so there was little mismatch between the TTS speech and the underlying ASR model. One can imagine that the speech encoder portion of the model removes speaker variability, etc. to produce an embedding that depends largely on just the word sequence; hence, any representative TTS speech would be sufficient because the weakest link was the intent classifier. Finally, if we combine the two methods, joint training text and speech embeddings with synthetic TTS speech data, we obtain modest gains, resulting in an intent classification accuracy of 88.3\%. 

\section{Conclusion}
\label{sec:conclusion}

End-to-end spoken language understanding systems require paired speech and semantic annotation data, which is typically quite scarce compared to NLU resources (semantically annotated text without speech).  We have made progress on this important but neglected issue by showing that an end-to-end speech-to-intent model can learn from annotated text data without paired speech.  By leveraging pre-trained text embeddings and data augmentation using speech synthesis, we are able to improve the intent classification error rate by over 60\% and achieve over 80\% of the improvement from paired speech-to-intent data. 

\section{Acknowledgements}
We thank Ellen Kislal for her initial studies on E2E S2I.

\vfill\pagebreak

\bibliographystyle{IEEEbib}
\bibliography{refs}

\end{document}